\documentclass[manuscript]{acmart}

\usepackage[figurename=Figure.]{caption}
\usepackage{cite}
\usepackage{amsmath,amssymb,amsfonts}
\usepackage{graphicx}
\usepackage{textcomp}
\usepackage{xcolor}
\usepackage{algpseudocode}
\usepackage{algorithm}
\usepackage{commath}
\usepackage[symbol]{footmisc}
\copyrightyear{2025}
\acmYear{2025}
\setcopyright{rightsretained}
\acmConference[CSAIDE 2025]{2025 4th International Conference on Cyber Security, Artificial Intelligence and the Digital Economy}{March 07--09, 2025}{Kuala Lumpur, Malaysia}
\acmBooktitle{2025 4th International Conference on Cyber Security, Artificial Intelligence and the Digital Economy (CSAIDE 2025), March 07--09, 2025, Kuala Lumpur, Malaysia}
\acmPrice{}
\acmDOI{10.1145/3729706.3729817}
\acmISBN{979-8-4007-1271-5/25/03}
\begin{document}

\title{Survey of Swarm Intelligence Approaches to Search Documents Based On Semantic Similarity}

\author{Chandrashekar Muniyappa}
\authornotemark[1]
\affiliation{%
  \institution{School of EECS, College of Engineering and Mines, University of North Dakota}
  \city{Grand Forks}
  \country{USA}}
\email{c.muniyappa@und.edu}

\author{Eunjin Kim}
\affiliation{%
  \institution{School of EECS, College of Engineering and Mines, University of North Dakota}
  \city{Grand Forks}
  \country{USA}}
\email{ejkim@und.edu}

\renewcommand{\shortauthors}{Muniyappa et al.}

\begin{abstract}
Swarm Intelligence (SI) is gaining a lot of popularity in artificial intelligence, where the natural behavior of animals and insects is observed and translated into computer algorithms called swarm computing to solve real-world problems. Due to their effectiveness, they are applied in solving various computer optimization  problems. This survey will review all the latest developments in Searching for documents based on semantic similarity using Swarm Intelligence algorithms and recommend future research directions.
\end{abstract}


\maketitle

\section{Introduction}
\begin{figure*}
    \centering
    \includegraphics[width=0.8\textwidth]{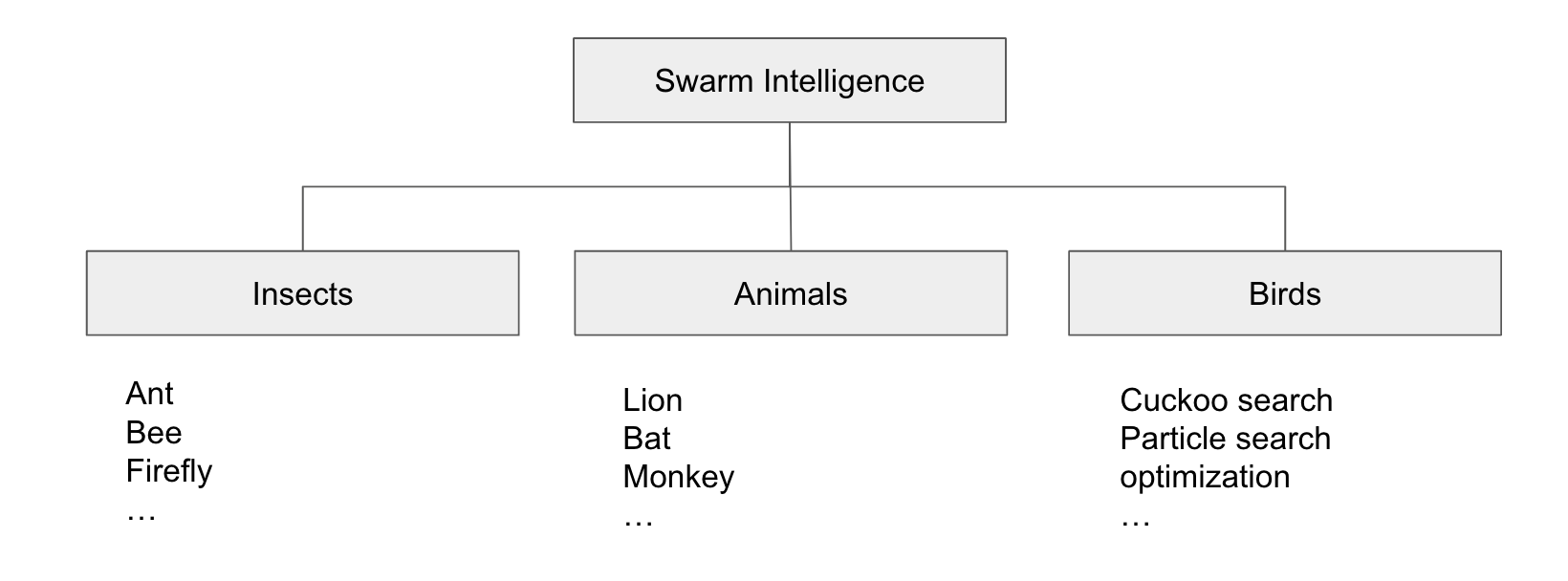}
    \caption{Sample of Swarm Intelligence Hierarchy}
    \label{fig:img1}
\end{figure*}

A swarm refers to a group of agents that communicate with each other directly or indirectly using their local environment to solve a problem. Computer scientists studied the group behavior of animals and insects and developed computer algorithms to solve complex real-world problems which led to Swarm computing. However, with cloud computing volume of data is growing exponentially, to build complex systems large volumes of data have to be explored. Therefore, deterministic systems will not be able to scale, to solve this problem, meta-heuristics algorithms are used. Swarm Intelligence (SI) meta-heuristic algorithms that can learn and adapt to dynamically changing environments like biological organisms are gaining much popularity in solving complex problems. They are effective in searching large solution spaces, and multi-dimensional hyper-planes, and adapting to dynamically changing constraints. There are different categories of SI algorithms, a sample of categories are shown in Fig \ref{fig:img1}. In this paper, we will review how Particle Swarm Optimization (PSO), Ant Colony Optimization (ACO), and variants of these algorithms are applied to various text similarity measurement problems.

\section{Main}
X.-F. Song et al. [1], applied particle swarm optimization to solve feature selection problem in high-dimensional dataset. The approach has three steps:
In the first step Symmetric Uncertainty (SU) method was applied to compute the relevancy score and the values less than the predetermined threshold will be elemenated as irrelevant features. In the second step, features were clustered using the SU score computed in the first step; such that, if the difference bewteen their SU score is high they belong to different clusters; otherwise, to the same cluster. Finally in the third step, PSO algorithm is applied to pick the representative features from each cluster 
as the final answer. To apply PSO, the particles were encoded using the integer values of cluster number they belong to, for example if a feature $\mathit{F_i}$ belongs to cluster $\mathit{C_j}$, then it was encoded using the integer value $j$. As all values are integers, the integer programming PSO a variant of PSO [2] was applied. The optimization process was guided using the relevancy of the given feature with the cluster. The relevancy probability was computed using equation: \ref{relevancyPSO}

\begin{equation}\label{relevancyPSO}
pc_j=\frac{Cv_j^{max}}{Cv}
\end{equation}

where
\begin{itemize}
\item $j$ is the number os clusters 1,2,3,...M. 
\item $Cv_j^{max}$ is the max relevancy score of the given cluster $j$
\item $Cv$  is the relevancy score of the given feature for the given cluster
\item $pc_j$ is the relevancy probability of the given particle
\end{itemize} 

The process is repeated for all the particles, for the specified number of iterations using classification accuracy as the fitness function. During experimentation they found that when $\mathit{p_{best}}$ is near equal to $\mathit{g_{best}}$ that is when a particle's local best is ner equal to global best then the algorithm was stuck in the local minima resulting in poor results. To overcome this problem and to increase the diversity of the particles selection, a novel difference mutation strategy to randomly initialize the particles in the generations was applied.
\par
J. Gao et al [3] modified the velocity term in the traditional PSO to probabilities and suggests the Clustering Probabilistic Particle Swarm Optimization (CPPSO) algorithm for feature selection. Instead of using binary values as the velocity to include or exclude the feature; floating point probabilities are used to select the features. The probabilities are calculated using the equation \ref{cppso}

 \begin{align}
  \begin{aligned}\label{cppso}
p_{i,j}^{t+1} = p_{i,j}^{t}= w * p_{i,j}^{t} + c_1 * r*(P_{best} -x_{i,j}^t)\\
    +c_2 * r*(G_{best} -x_{i,j}^t)- r*(D_r-x_{i,j}^t)
  \end{aligned}
 \end{align}

\begin{align*}
\left\{
    \begin {aligned}
         & x_{i,j}^{t+1} = 1- x_{i,j}^{t}  \quad & p_{i,j}^{t+1} \geq r \\
         & x_{i,j}^{t+1} = x_{i,j}^{t} \quad & otherwise                  
    \end{aligned}
\right.
\end{align*}

where
\begin{itemize}
\item ${p_{i,j}^{t}}$ is the probability of ${i^{th}}$ particle in ${j^{th}}$ dimension in iteration $t$.
\item $P_{best}$ and $G_{best}$  are the local and global optima
\item $D_r$  is the set of random numbers
\end{itemize} 

Random numbers helps in exploration and elitism is used for exploitation. These together boost the algorithm performance. K-means clustering algorithm [12] with Hamming distance is used to measure how dissimilar the particles are. The particles close to centroid are grouped together and the rest is grouped as unfavorable solutions. The steps are repeated for the specified iterations and finally the cluster with small radius is picked as the solution.

\par
S. Rohaidah Ahmada et al [4], deviced ACO-KNN an implementation of KNN algorithm using the Ant Colony Optimization (ACO), a hybrid algorithm to extract features from text for sentiment classification. They split the process into two phases. In the first phase, they cleaned the text by removing stopwords, stemming, and by applying other Natural Language Programming (NLP) cleaning tasks [10]. And then clustered the documents using Fuzzy C-Means (FCM) clustering algorithm [5]. In the second phase, the extracted features were represented as a graph and ants were placed randomly on each node to start the search. Each ant kept track of local best and global best, where local best is the best solution found by an Ant and global best refers to the optimal solution found based on the work done by all ants. At the end of each iteration, KNN algorithm is applied to classify text sentiments with the subset of features found by ants and Mean Squared Error (MSE) is calcualted, lower the MSE value better is the result. Based on the results Pheromone table is updated and Ant movements are decided. The process is repeated for the specified number of iterations or until there is no improvement in the MSE. They tested this hybrid algorithm by applying it on opensource datasets and compared the results with Genetic Algorithm (GA), the results obtained by ACO-KNN were far superior. The equations \eqref{lant} gives the pheromone updation rule for local best when an ant completes the route.

\begin{equation}\label{lant}
\Delta\tau_i^k(t)= \phi \cdot \gamma(S^k(t)) + \frac{w\cdot(n-|{S^k(t)|})}{n} 
 if i \in S^k(t)
\end{equation}
Where:
\begin{itemize}
\item $n$ is the number of features
\item $(S^k(t))$ is the subset of features found by ant k at iteration t
\item $|{S^k(t)}|$ length of feature subset found by ant k at iteration t
\item $\gamma(S^k(t))$ classification performance
\item $\phi$ relative weight of classifier performance (0.8)
\item $w$ relative weight of feature subset length (0.2)
\item $\phi \in [0,1]$
\item $w$ = 1 - $w$
\end{itemize}

Once all the ants have completed the route, equation \eqref{gant} is used to update the global best route.

\begin{equation}\label{gant}
\Delta\tau_i^g(t)= \phi \cdot \gamma(S^g(t)) + \frac{w\cdot(n-|{S^g(t)|})}{n} if i \in S^g(t)
\end{equation}
Where:
\begin{itemize}
\item $n$ is the number of features
\item $(S^g(t))$ is the global best subset of features found in iteration t.
\item $|{S^g(t)}|$ length of the global best feature subset found in iteration t.
\item $\gamma(S^g(t))$ classification performance
\item $\phi$ relative weight of classifier performance (0.8)
\item $w$ relative weight of feature subset length (0.2)
\item $\phi \in [0,1]$
\item $w$ = 1 - $w$
\end{itemize}

\par
F. Rehaman et al [6] applied Ant Colony Optimization (ACO) to build a diet-recommendation system based on user health problems. To design the system, the initial data setup  like pathology tests, food items, and ingredients of each food item was manually set up. When a user entered the health issues to seek food recommendations, the system matched input with pathology test results and pulled the food items list based on doctor-recommended ingredients. However, the list of items must be fine-tuned to match the user's needs. The food items are represented as graph nodes and edges are defined by heuristic and pheromone weights to help ants determine the optimal results. A random set of Ants are placed on a few nodes and the search process is started.  The local best solution is updated with the equation \eqref{ldiet}.

\begin{equation}\label{ldiet}
\tau_\textbf{ij}^k(t+1)= \begin{cases}
u^k, & \text{if $\textit{ij} \in S^k(t)$}\\
            (1-\rho) \tau_\textit{ij}^k(t) & \text{otherwise}
		 \end{cases}
\end{equation}

Where:
\begin{itemize}
\item $\tau_\textbf{ij}^k(t+1)$ pheromone level increase by $\delta\tau^k$
\item $\rho$ is the pheromone decay parameter, $k$ represents ant, $t$ represents the iteration
\end{itemize}

For the global optimal solution the same equation \eqref{ldiet} is used, but solution $k$ from a specific ant will be replaced by $g$ the best solution based on the work done by all ants. Heuristic value $\eta$ is used to control exploration and exploitation. The problem is modeled as supervised learning and accuracy is measured using Root Mean Squared Error (RMSE) to determine the ant movements. Through experiments, they prove results obtained are good. 

\par
M. Yogi et al [7] combined Neighborhood Preserving Embedding (NPE) and Particle Swarm Optimization (NPE-PSO) algorithms and devised a hybrid approach to classify web documents. Instead of using standard text similarity measurement techniques like Euclidean, Cosine [14], Jaccard similarity, and so on, they used NPE which will maintain the local neighborhood information while reducing the dimensionality of input features and has a better performance when compared to Principal Component Analysis (PCA) [8]. The web documents were fed into NPE to reduce them into feature vectors, these feature vectors were used as swarm particles for the PSO algorithm. The evolution process starts with a user search keywords as the initial population and then evolve the vectors to get more relevant particles. The process is repeated for the specified number of iterations and the results are returned. The implementation is compared with text classification using Ant Colony Optimization (ACO), and NPE-ACO. The results obtained by NPE-PSO were better when compared to other implementations. This approach uses vector representation of features that can capture similarity better than other representations and uses an advanced algorithm like NPE to measure the similarity.

\par
D. Yang et al [9] applied Ant Colony Optimization (ACO) algorithm to cluster athletes based on their behavior. In this approach they use only ACO algorithm without integrating it with any other algorithm and through experiments prove that the results good compared to other clustering algorithms. Before applying the algorithm they applied different NLP [10] techniques to clean the data and reduce the noise. Following steps were followed to cluster the data. The pheromone level corresponding to each feature is randomly initialized, the position of ants are randomly assigned to different features. The ants select the next feature based on the pheromone level and distance. Features with higher pheromone level and closer would be the best next step. At the end of each step pheromone level is updated using the equation \ref{phero} if the level of pheromone increases the solution is good; bad, otherwise. Finally the algorithm stops when the specified number of iterations are done or when the pheromone change reaches certain threshold.

\begin{equation}\label{phero}
\tau_{ij} = (1-\rho) * \tau_{ij} + \sum_{k=1}^{m} \Delta\tau_{ij}^k
\end{equation}

where
\begin{itemize}
\item $\tau_{ij}$ is the pheromone level between i and j data points.
\item $\rho$ is the pheromone decay parameter, $k$ represents ant.
\item $\Delta\tau_{ij}^k$ is the increment in pheromone level by the k-the ant.
\end{itemize}

The idea here is Ants attract other Ants in a specific path increasing the pheromone level as long as the solution is desirable. In clustering it will select feature that consistently contribute to increase in pheromone level there by grouping them into the same cluster. The purity of the luster was measured using the Davies-Bouldin (DB) index [11], which measures the compactness within the cluster and the separation between the clusters as shown in equation \ref{dbindex}.

\begin{equation}\label{dbindex}
DB =1/n \sum_{i=1}^{n} max_{j\neq i} \left( \frac{\sigma_i +\sigma_j}{d(c_i,c_j)} \right)
\end{equation}

where
\begin{itemize}
\item $\tau_{i}$ $\tau_{i}$ are the distance of a data point to the center of the clusters i and j. 
\item $d(c_i,c_j)$ is the distance between the cluster centers i and j
\end{itemize}

The lower the value of Davies-Bouldin (DB) index, the better the results.

\par
S. Gite et al [15] applied Ant Colony Optimization (ACO) algorithm to extract text features to classify hate speech. As part of this research, they applied Machine Learning (ML) models like Logistic Regression, Random Forest, K Nearest Neighbours on the features extracted using ACO to improve the classification accuracy. The text data was collected from Twitter and then NLP text cleaning steps were followed to extract the keywords and then TF-IDF and Bag-of-words approaches were used to generate the embedding vectors. The problem was modeled as a weighted graph, where the nodes were represented by the feature vectors and the edges were represented by the heuristic parameters to control the ant movements. The algorithm was repeated for the user-specified iterations and then finally the subset of features selected by ACO was used as the features to train the aforementioned ML classification models. The models were tested with and without features selected from ACO. The results showed that the classification accuracy improved by $10\%$ with ACO extracted feature subset. This research shows, how evolutionary and adaptive meta-heuristic algorithms can be used to improve the performance of traditional ML models. 

\par
P. Moradi et al [16] devised a novel graph clustering with ant colony optimization for feature selection (GCACO) algorithm. As part of this research, they developed a novel graph clustering approach to select the optimum number of features by applying ACO. This approach was devised to resolve the following drawbacks of the regular ACO. It is standard practice to represent the problem as a fully connected graph for the ACO search problem. However, this will have 
$n!/(n-m)!$ time complexity, where $n$ is the number of nodes and $m$ is the number of edges that can be further reduced. To update the pheromone, usually ML models are used to measure the accuracy at the end of each iteration, which will add to the time complexity. Besides, ACO does not keep track of previous solutions, based on the ML model accuracy it will pick the subset of features which may have redundant features. Finally, the solution set size will be controlled by the ants' travel path which is usually a constant parameter to the algorithms, therefore, multiple iterations are required to find the right value. To solve all these problems, following three steps were followed. First, features from the dataset or text documents were extracted and represented as a weighted graph, where the weights represent the similarity between the features. Secondly, the graph was clustered using a community detection algorithm. Finally, the ACO algorithm was run on the clustered graph to find the optimal feature set. Let us take a detailed look at each of these steps.
\par
The input data can be structured or unstructured data. Once, we have the features they are represented as graph nodes. The edges are added to build a fully connected graph and weights are assigned to each edge based on the similarity between the source and destination features (nodes) of each edge. The similarity score is computed using the Pearson correlation given in the equation \eqref{pearson}.

\begin{equation}\label{pearson}
W_ij=\abs{\frac{\sum_{p}(x_i-\Vec{x_i})(x_j-\Vec{x_j})}{\sqrt{\sum_{p}(x_i-\Vec{x_i})^2}\sqrt{\sum_{p}(x_j-\Vec{x_j})^2}}}
\end{equation}

Where:
\begin{itemize}
\item $W_ij$ is the similarity score between the nodes $i$ and $j$
\item $x_i$ and $x_j$ are the feature vectors and $\Vec{x_i}$ and $\Vec{x_j}$ are their respective mean values. 
\end{itemize}

Once the weighted graph is built, the next step is to cluster the graph. Traditional clustering algorithms have the following problems. Firstly, the number of clusters $k$ has to be specified; however, determining the right number of clusters is an exhaustive task. Therefore, the algorithm has to be experimented with different values of $k$. Secondly, distributing the data between different clusters is a challenging task. Lastly, every cluster will contribute equally to the final feature set. To overcome these problems, a community detection algorithm was used. Community detection algorithms will group the nodes that are highly correlated into the same community thereby clearly separating the dissimilar ones. Dynamically, determining the number of clusters based on the input data points. 

The last step is to apply the ACO algorithm, the ants are randomly placed on different community clusters and pheromone intensity is used to control the navigation of ants from one cluster to another. As the ants are moving they can either pick a feature within the cluster or from a different cluster. If a large number of features are picked from within a cluster then the final result will have correlated or redundant features. Therefore, the features should be picked from multiple clusters as much as possible to build an optimal feature set. Hence, ants will keep moving until they have visited all the clusters. The probability of an ant $k$ picking the next feature $F_j$ is calculated using the following equation \eqref{gcacopick}.

\begin{equation}\label{gcacopick}
P_k(F_j,VF_K)=\begin{cases}
		\frac{[\tau_j]^\alpha[\eta(F_j,VF_k)]^\beta}{\sum_{u \in UF_i^k}[\tau_u]^\alpha[\eta(F_u,VF_k)]^\beta},&\text{if } j \in UF_i^k  \\
            0, & \text{Otherwise}
		 \end{cases}
\end{equation}

Where:
\begin{itemize}
\item $\tau_j,\tau_u$ is the pheromone intensity of features $j , u$.
\item $\eta(F_j,VF_k)$ represents the heuristic information function, along with constants $q , q_0$ control the exploration and exploitation.
\item $UF_i$ is the set of unvisited features by ant $k$
\item $ \alpha, \beta$ are parameters to determine the importance of pheromone level versus the heuristic values.
\end{itemize}

Each step is repeated until every ant has visited all the clusters. At the end of each iteration pheromone level of features is updated and the process is repeated for the user-specified number of iterations. The pheromone update rule is given in equation \eqref{gcacopheromone}.

\begin{equation}\label{gcacopheromone}
\tau_i(t+1)=(1- \rho)\tau_i(t) + \sum_{k=1}^{A} \Delta_i^k(t)    
\end{equation}

Where:
\begin{itemize}
    \item $\tau_i(t) , \tau_i(t+1)$ are the amount of pheromone at step $t$ and $t+1$
    \item $\rho$ is the pheromone decay parameter.
    \item $\Delta_i^k(t)$ is the extra pheormone for the feature $F_i$ by the ant $k$.
\end{itemize}

They applied the algorithm on multiple opensource datasets and compared the results by using the selected features with multiple ML models and found that the results obtained were far superior when compared to other feature selection techniques. The features obtained by GCACO algorithm where around 23; however, for the same datasets features obtained by applying other techniques were much larger. This clearly proves the efficiency of this approach. 

T. Londt et al [17] applied Multi-Objective Particle Swarm Optimization algorithm to extract the features from text to build classification models. They used the Reuters R8 news dataset for this research, as the first step they applied NLP text cleaning steps [10] to clean the dataset and extract the features based on the TF-IDF technique [13]. There are different ways of selecting features. In the filter-based feature selection technique, features  are ranked based on their importance in explaining the data behavior. Usually, they are simple and fast in computing; however, they may not capture all the features. Therefore, the accuracy of the final model will be low. To overcome this problem, a wrapper-based technique was used, where advanced evolutionary algorithms were used to select the important features and accuracy was measured using the classification model and the results were used to guide the search space. Hence, the overall accuracy was good, but computationally expensive.

Based on these observations, they proposed a Two-stage multi-objective PSO algorithm for feature selection. In the first stage, they used well know Information Gain (IG) [20], Gain Ratio (GR) [21], correlation (CO) [22], and Symmetrical Uncertainty (SU) [23] filter-based feature selection techniques to select the most important features. In the second stage, they applied a multi-objective PSO algorithm to select the optimal feature set by wrapping the Naive Bayes ML classification model. The first objective function is shown in the equation \eqref{objective1}. Known as the balanced accuracy Objective function. 

\begin{equation}\label{objective1}
    Objective_1(\Vec{x})=\frac{1}{c} \sum_{n=1}^{c}\frac{TP_i}{|S_i|}
\end{equation}

Where:
\begin{itemize}
    \item $c$ is the number of classes in the given dataset.
    \item $TP_i$ is the True Positive rate, the number of items correctly classified for the class $i$.
    \item $|S_i|$ is the total number of data points in class i.    
    \item $\frac{1}{c}$ is used to provide equal weightage to all the classes.
\end{itemize}

In the real world not all the datasets will be balanced, meaning not all the classes or categories of data points will be of equal number in the dataset. When we have unbalanced datasets, the ML model will result in over-fitting, meaning wrongly classifying minority classes into the majority category. To solve this problem equation \eqref{objective1} was used as the first minimization function. The second objective function is given in the equation \eqref{minimize} to minimize the number of features selected by the algorithm.

\begin{equation}\label{minimize}
    Objective_2(\Vec{x})= |\Vec{x}|
\end{equation}

Where:
\begin{itemize}
    \item $\Vec{x}$ is the feature vector selected by the algorithm.
\end{itemize}

The main idea was to achieve maximum classification accuracy by selecting the minimum number of optimal features. In every iteration, the features selected
in the first stage was fed as the input to the second stage where the evolutionary algorithm picked the best set of features by applying the above-mentioned objective functions based on crowding, mutation, and dominance concepts [18]. Crowding refers to picking the winning features based on the tournament selection and adding them to the leader set. A mutation is a process of diversifying the feature swarm, so that, distinct features are selected in the final result set. And dominance refers to fixing the size of the result set to be returned by the algorithm. Hamming distance [19] was used to measure the text feature similarity as part of searching the solution space.
At the end of each iteration, the best solutions are recorded and the Naive Bayes ML model was used to measure the accuracy of the features selected. The process was repeated for the specified number of iterations or until there was no improvement in accuracy. The algorithm was tested on the Reuters R8 dataset and the performance was compared with two-single objective PSOs with the aforementioned objective functions. Furthermore, the results were compared with other feature selection techniques. The results produced by the proposed approach produced distinct sets of a minimum number of features whose classification accuracy was better than the compared algorithms.
\par
\section{Conclusion anf Future Work}
Based on these studies, we can observe that researchers have used different types of text representation techniques like keyword-based TF-IDF, vector-based embeddings, and different types of graph representations. And applied PSO, ACO, variants of both, and hybrid Swarm Intelligence algorithms to solve different types of clustering, recommendation, and feature extraction problems. Each study consistently showed that traditional text similarity measurement techniques can be improved by applying Swarm Intelligent algorithms. Based on this survey we recommend the following future work. The research work to address the following observations is in-progress.

\begin{description}
    \item[$\bullet$] None of them applied Swarm Intelligence algorithms like PSO (Particle Swarm Optimization) and ACO (Ant Colony Optimization) on advanced sentence embedding vectors text representations  to identify documents based on semantic similarity.
    \item[$\bullet$] Most of the techniques used traditional similarity measurement techniques such as cosine similarity, Euclidean distance, Dice distance, and many more. They all work well on short texts; however, we need to evaluate their performance on large paragraphs of texts along with the swarm intelligent algorithms.
    \item[$\bullet$] When the size of text increases, the length of embedding vectors also increases to capture the semantic similarity between texts effectively. We need to evaluate the performance of swarm intelligence algorithms on these large dimensional solution spaces and identify the type of operations that can be applied on these embedding vectors that help the algorithms to converge faster. 
    
\end{description}

\end{document}